\documentclass[conference]{IEEEtran}
\IEEEoverridecommandlockouts
\usepackage{cite}
\usepackage{amsmath,amssymb,amsfonts}
\usepackage{algorithmic}
\usepackage{graphicx}
\usepackage{textcomp}
\usepackage{xcolor}
\usepackage{booktabs}
\usepackage{url}
\def\BibTeX{{\rm B\kern-.05em{\sc i\kern-.025em b}\kern-.08em
    T\kern-.1667em\lower.7ex\hbox{E}\kern-.125emX}}
\begin{document}

\title{Diffusion Models with Anisotropic Gaussian Splatting for Image Inpainting}

\author{\IEEEauthorblockN{Jacob Fein-Ashley}
\IEEEauthorblockA{
\textit{University of Southern California} \\
\texttt{feinashl@usc.edu}}

\and

\IEEEauthorblockN{Benjamin Fein-Ashley}
\IEEEauthorblockA{
\textit{University of Southern California} \\
\texttt{jfeinash@usc.edu}}


}

\maketitle

\begin{abstract}
Image inpainting is a fundamental task in computer vision, aiming to restore missing or corrupted regions in images realistically. While recent deep learning approaches have significantly advanced the state-of-the-art, challenges remain in maintaining structural continuity and generating coherent textures, particularly in large missing areas. Diffusion models have shown promise in generating high-fidelity images but often lack the structural guidance necessary for realistic inpainting.

We propose a novel inpainting method that combines diffusion models with anisotropic Gaussian splatting to capture both local structures and global context effectively. By modeling missing regions using anisotropic Gaussian functions that adapt to local image gradients, our approach provides structural guidance to the diffusion-based inpainting network. The Gaussian splat maps are integrated into the diffusion process, enhancing the model's ability to generate high-fidelity and structurally coherent inpainting results. Extensive experiments demonstrate that our method outperforms state-of-the-art techniques, producing visually plausible results with enhanced structural integrity and texture realism.
\end{abstract}

\section{Introduction}

Image inpainting involves filling in missing or corrupted regions of an image in a manner that is visually indistinguishable from the original content. This has wide applications in photo editing, object removal, image restoration, and occlusion handling in vision systems~\cite{bertalmio2000image, criminisi2004region}.

Traditional inpainting techniques can be broadly categorized into diffusion-based and exemplar-based methods. Diffusion-based methods~\cite{bertalmio2000image, ballester2001filling} propagate pixel information from the known regions into the missing areas using partial differential equations. While effective for small holes and smooth textures, these methods often produce blurred results and fail to reconstruct complex structures.

Exemplar-based methods~\cite{criminisi2004region} fill missing regions by sampling and copying patches from the known parts of the image. These approaches better preserve texture details but struggle with structural coherence, especially when suitable patches are not readily available in the source image.

With the advent of deep learning, convolutional neural networks (CNNs) have been employed to learn semantic features for inpainting. Pathak et al.~\cite{pathak2016context} introduced Context Encoders that leverage an autoencoder architecture to predict missing content. Iizuka et al.~\cite{iizuka2017globally} improved upon this by ensuring global and local consistency using separate discriminators.

Attention mechanisms have further enhanced inpainting performance by allowing models to reference distant contexts within the image. Yu et al.~\cite{yu2018generative} proposed a Contextual Attention module enabling the network to utilize relevant information from distant spatial locations. Liu et al.~\cite{liu2018image} addressed irregular missing regions using partial convolutions, and Yu et al.~\cite{yu2019free} introduced gated convolutions with learnable dynamic feature selection.

\subsection{Motivation}

Despite these advancements, generating high-quality inpainting results remains challenging, particularly for images with large missing regions and complex structures. Existing methods may struggle to maintain structural continuity and produce visually coherent textures. One fundamental difficulty lies in effectively capturing both the local structures (edges and textures) and the global context (such as object coherence and scene semantics) required for realistic inpainting.

Recently, \textbf{diffusion models} have emerged as a powerful class of generative models capable of producing high-quality images~\cite{sohl2015deep, ho2020denoising, song2020score}. Diffusion models define a forward process that gradually adds noise to the data and a reverse process that removes the noise to recover the data distribution. In image inpainting, diffusion models offer advantages in generating diverse and high-fidelity content, as they can model complex data distributions and avoid issues like mode collapse encountered in GANs.

However, diffusion models in their basic form may not effectively leverage structural priors inherent in images, such as edges and textures, which are crucial for maintaining structural continuity in inpainting tasks. This motivates the integration of structural guidance into the diffusion process.

On the other hand, \textbf{Gaussian splatting} has been employed in neural rendering to represent scenes using oriented Gaussian functions, capturing spatial influence and uncertainty~\cite{gao2023surfelnerfneuralsurfelradiance}. Gaussian functions can model the spatial influence of missing pixels in two-dimensional image domains, guided by local image structures, providing a probabilistic framework for representing uncertainty and structural cues.

Integrating anisotropic Gaussian functions into the inpainting process allows us to encode the local geometry and texture information, guiding the inpainting model to focus on essential regions and maintain structural coherence. By adapting the covariance of the Gaussian functions based on local gradients and incorporating multi-scale information, we can capture both fine details and global context.

\subsection{Our Approach}

Motivated by these observations, we propose a novel inpainting method that combines diffusion models with anisotropic Gaussian splatting to capture local structures and global context effectively. Our approach leverages the strengths of diffusion models in generating high-fidelity content and integrates structural guidance through Gaussian splatting to maintain structural continuity.

Specifically, we model the missing regions using anisotropic Gaussian functions that adapt to the local gradient information, capturing the spatial influence and uncertainty around missing pixels. We compute Gaussian splat maps at multiple scales to capture information at different resolutions. These Gaussian splat maps serve as guidance for a diffusion-based inpainting network, informing the model about the image's structural and spatial priors.

By integrating the Gaussian splat maps into the diffusion process, our model benefits from both the generative capabilities of diffusion models and the structural guidance provided by the Gaussian functions. This synergistic combination allows our method to generate inpainting results that are visually coherent and structurally consistent with the known regions of the image.

\subsection{Our Contributions}

This work presents a novel inpainting approach that incorporates anisotropic Gaussian splatting into a diffusion-based generative model. Our main contributions are:

\begin{itemize}
    \item \textbf{Anisotropic Gaussian Modeling}: We introduce anisotropic Gaussian functions that adapt to the local gradients of the image, providing a more accurate representation of spatial influence than isotropic models. This modeling captures the uncertainty and structural cues around missing regions.
    \item \textbf{Integration with Diffusion Models}: We incorporate the Gaussian splat maps into a diffusion-based inpainting network, guiding the diffusion process with structural priors. This integration enhances the model's ability to generate high-fidelity and structurally coherent inpainting results.
    \item \textbf{Multi-Scale Gaussian Splatting}: By computing Gaussian splat maps at multiple scales, we capture fine details and larger contextual information, improving the network's understanding of the image structure at different resolutions.
    \item \textbf{Comprehensive Evaluation}: We conduct extensive experiments comparing our method with state-of-the-art inpainting algorithms, demonstrating superior performance in quantitative metrics and visual quality.
\end{itemize}


\section{Related Works}

Image inpainting aims to restore missing or corrupted regions of an image in a visually plausible way. Traditional methods are generally divided into diffusion-based and patch-based techniques, while more recent approaches leverage deep learning, including generative adversarial networks (GANs) and diffusion models.

\subsection{Traditional Inpainting Methods}

\subsubsection{Diffusion-Based Methods}

Early inpainting algorithms employ diffusion processes to propagate information from known regions into missing areas. Bertalmio et al.~\cite{bertalmio2000image} introduced a technique that mimics the manual inpainting process used by artists, using partial differential equations to fill in the missing regions by propagating linear structures called isophotes. Ballester et al.~\cite{ballester2001filling} extended this work by relating it to the Mumford-Shah functional, improving the capability to preserve edges and smoothness.

However, traditional diffusion-based methods tend to produce blurred results when dealing with large missing regions or complex textures, as they lack mechanisms to reproduce high-frequency details.

\subsubsection{Patch-Based Methods}

Patch-based methods address the limitations of diffusion approaches by sampling and copying patches from known regions to fill the missing areas. Efros and Leung~\cite{efros1999texture} proposed a non-parametric texture synthesis method that forms the basis for many exemplar-based inpainting algorithms. Criminisi et al.~\cite{criminisi2004region} introduced a prioritized patch-filling technique that considers both texture synthesis and edge propagation, resulting in improved structural continuity.

Despite their effectiveness in preserving local textures, patch-based methods can struggle with global coherence and may produce artifacts when suitable patches are not available.

\subsection{Deep Learning-Based Inpainting}

\subsubsection{Context Encoders and GANs}

The advent of deep learning brought significant advancements in image inpainting. Pathak et al.~\cite{pathak2016context} introduced Context Encoders, using convolutional neural networks (CNNs) to predict missing content by learning from large datasets. They employed an adversarial loss to encourage the generator to produce realistic images.

Building upon this, Iizuka et al.~\cite{iizuka2017globally} proposed a generative model that ensures both global and local consistency, using two discriminators to capture features at different scales.

\subsubsection{Attention Mechanisms}

Attention mechanisms have been incorporated to improve the inpainting of complex structures. Yu et al.~\cite{yu2018generative} developed a Contextual Attention module that allows the network to borrow relevant features from distant spatial locations, effectively capturing long-range dependencies.

Nazeri et al.~\cite{nazeri2019edgeconnect} introduced EdgeConnect, which uses edge detection as explicit structural guidance for image inpainting. The method consists of an edge generator and an image completion network, resulting in sharper and more coherent results.

\subsubsection{Partial and Gated Convolutions}

Liu et al.~\cite{liu2018image} proposed Partial Convolutional layers that handle irregular masks by normalizing the convolution operation based on the valid pixels. This approach allows the network to be more robust to varied hole patterns.

Yu et al.~\cite{yu2019free} extended this idea with Gated Convolutions, where the gating mechanism is learned dynamically, providing the network with the ability to determine the importance of features during the inpainting process.

\subsubsection{Diffusion Models for Image Inpainting}

Recently, diffusion models have emerged as a powerful class of generative models capable of producing high-quality images~\cite{sohl2015deep,ho2020denoising}. Diffusion models define a forward process that gradually adds noise to the data and a reverse process that removes the noise to recover the data distribution.

In the context of image inpainting, Song et al.~\cite{song2020score} introduced Score-Based Generative Models that use stochastic differential equations for image generation and inpainting. Saharia et al.~\cite{saharia2022palette} proposed Palette, a diffusion-based model for image editing tasks, including inpainting, which demonstrated superior performance in maintaining image fidelity and consistency. Lugmayr et al.~\cite{lugmayr2022repaint} introduced RePaint, adapting diffusion models for conditional image generation, showing impressive results in inpainting tasks by iteratively refining the missing regions through the reverse diffusion process.

Compared to GAN-based methods, diffusion models offer advantages in generating diverse and high-fidelity images without mode collapse. However, diffusion models generally require longer inference times due to the iterative denoising steps.

Our work incorporates a diffusion-based inpainting network conditioned on anisotropic Gaussian splatting, combining the strengths of diffusion models in generating realistic textures with the spatial guidance provided by Gaussian functions.

\subsection{Neural Rendering and Gaussian Splatting}

In the field of neural rendering, Gaussian splatting has been employed to represent 3D scenes~\cite{gao2023surfelnerfneuralsurfelradiance}. These methods model scenes using oriented 3D Gaussians, enabling efficient rendering and novel view synthesis.

Our work adapts Gaussian splatting to 2D image inpainting, introducing anisotropic Gaussian functions to model the spatial influence of missing pixels. By integrating this with a diffusion-based inpainting network, our method aims to capture both local structures and global context more effectively.

\subsection{Multi-Scale and Attention-Based Networks}

Multi-scale architectures have proven effective in capturing image features at different resolutions~\cite{ronneberger2015u}. Attention mechanisms further enhance the network's ability to focus on important regions~\cite{vaswani2017attention}.

Our approach combines these concepts by using a multi-scale network with attention guided by the Gaussian splat maps. Additionally, by integrating diffusion models within this framework, we leverage the strengths of both diffusion-based generation and attention mechanisms to produce high-quality inpainting results.

\section{Methods}

\begin{figure}[htbp]
\centering
\includegraphics[width=0.9\linewidth]{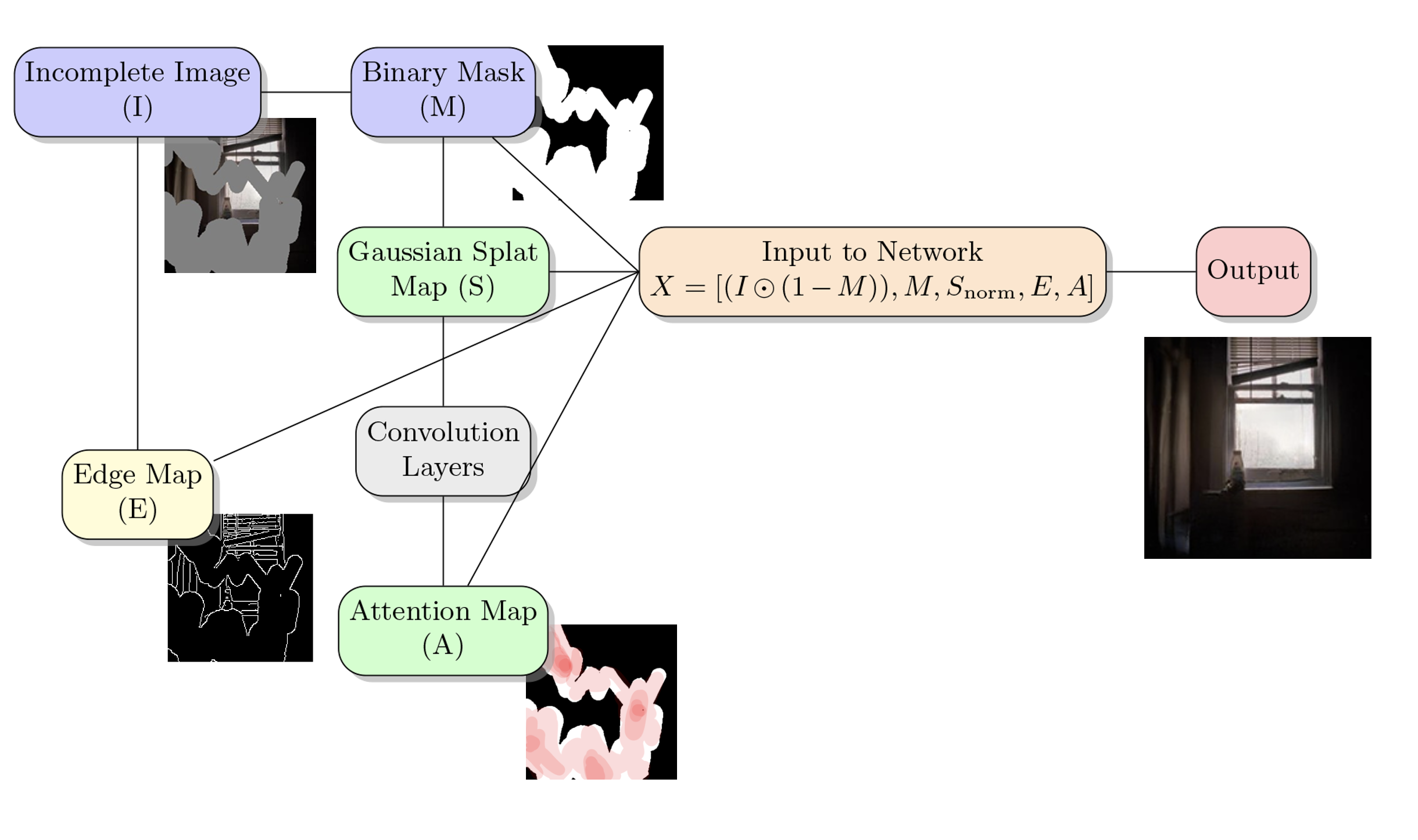}
\caption{Overview of the proposed inpainting method integrating anisotropic Gaussian splatting with a diffusion-based inpainting network. The anisotropic Gaussian splats model the spatial influence and uncertainty in the missing regions, guiding the diffusion process for more accurate restoration.}
\label{fig:model}
\end{figure}

In this section, we present a method for image inpainting and restoration using \textbf{diffusion models} in conjunction with \textbf{anisotropic Gaussian splatting}. The proposed method introduces anisotropic Gaussian modeling of missing regions, integration with a diffusion-based inpainting network, and a combination of loss functions for effective training. The method consists of three main components: enhanced Gaussian splatting of missing regions, a diffusion-based inpainting network guided by the Gaussian models, and the loss functions used for training. An overview of the proposed method is illustrated in Figure~\ref{fig:model}.

\subsection{Enhanced Gaussian Splat Modeling of Missing Regions}

\subsubsection{Anisotropic Gaussian Functions for Spatial Influence}

\begin{figure}[htbp]
\centering
\includegraphics[scale=0.3]{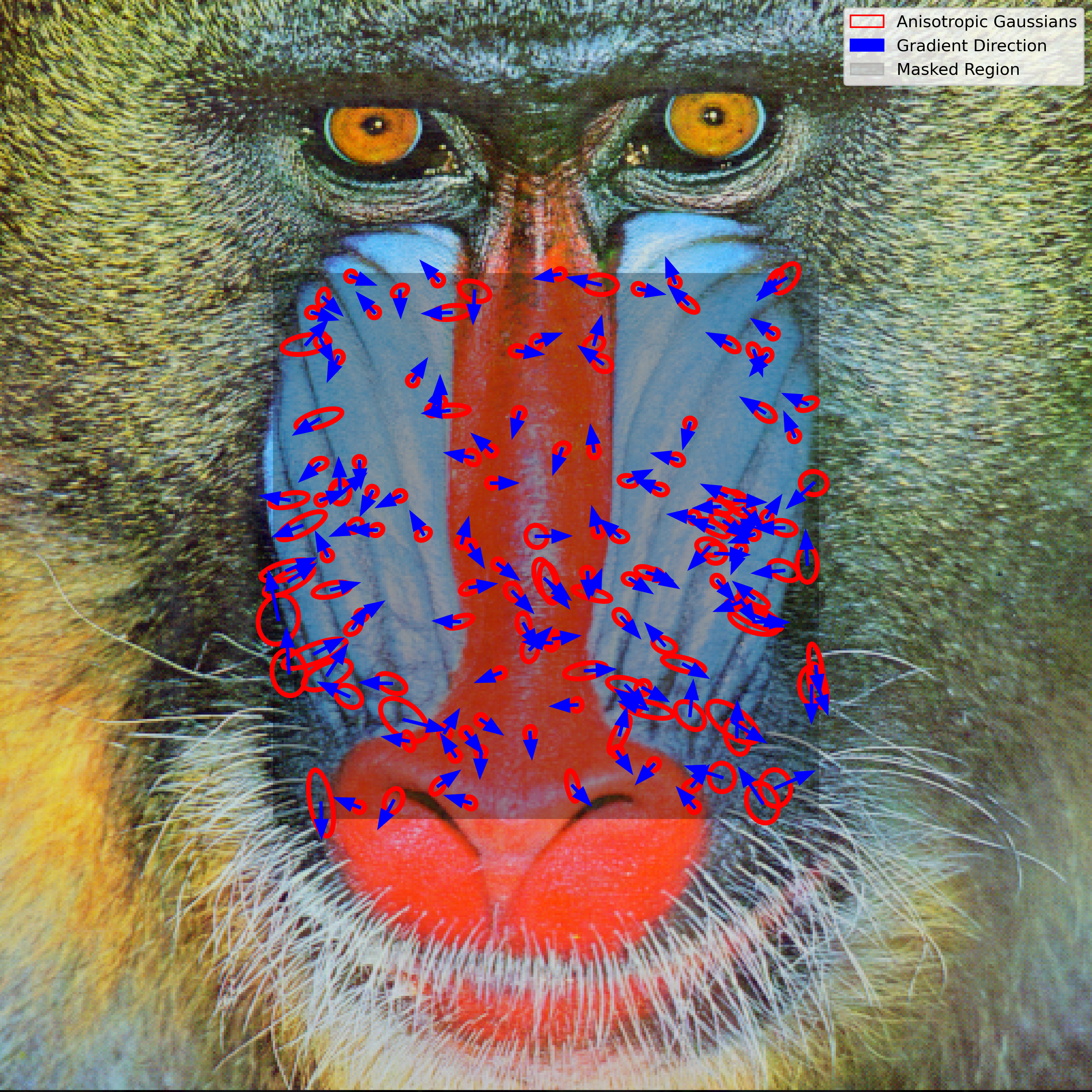}
\caption{Illustration of anisotropic Gaussian splatting in 2D, representing the spatial influence and uncertainty in missing regions. Each missing pixel is modeled as an anisotropic Gaussian function whose shape adapts to local image structures, allowing for more accurate guidance in the inpainting process.}
\label{fig:splatting2d}
\end{figure}

To capture the uncertainty and spatial influence around missing regions, we model each missing pixel using anisotropic Gaussian functions that adapt to local image structures. For each pixel location $(x, y)$ in the missing region ($M(x, y) = 1$), we define a 2D anisotropic Gaussian function $G_{x,y}$ as:

\begin{equation}
G_{x,y}(u, v) = \exp\left( -\frac{1}{2}
\begin{bmatrix}
u - x & v - y
\end{bmatrix}
\Sigma_{x,y}^{-1}
\begin{bmatrix}
u - x \\ v - y
\end{bmatrix}
\right)
\end{equation}

where $(u, v)$ are spatial coordinates in the image domain, and $\Sigma_{x,y}$ is the covariance matrix at pixel $(x, y)$, capturing the local structure and directionality~\cite{perona1990scale}. An illustration of the anisotropic Gaussian splatting is shown in Figure~\ref{fig:splatting2d}.

\subsubsection{Covariance Matrix Estimation}

We estimate the covariance matrix $\Sigma_{x,y}$ based on the local gradient information from the edge map $E$. Specifically, we compute the gradient vectors $\nabla E$ and estimate the structure tensor $\mathbf{J}_{x,y}$:

\begin{equation}
\mathbf{J}_{x,y} = 
\sum_{(i,j) \in \mathcal{N}(x,y)}
\begin{bmatrix}
E_x(i,j) E_x(i,j) & E_x(i,j) E_y(i,j) \\
E_x(i,j) E_y(i,j) & E_y(i,j) E_y(i,j)
\end{bmatrix}
\end{equation}

where $E_x$ and $E_y$ are the gradients of $E$ in the $x$ and $y$ directions, and $\mathcal{N}(x,y)$ is a local neighborhood around $(x,y)$~\cite{canny1986computational}.

We then define the covariance matrix $\Sigma_{x,y}$ as the inverse of the regularized structure tensor:

\begin{equation}
\Sigma_{x,y} = \left( \mathbf{J}_{x,y} + \epsilon \mathbf{I} \right)^{-1}
\end{equation}

where $\epsilon$ is a small positive constant to prevent singularity, and $\mathbf{I}$ is the identity matrix.

\subsubsection{Adaptive Amplitude Modulation}

To account for the varying influence of each Gaussian splat, we assign an amplitude $A_{x,y}$ based on the distance to the known regions:

\begin{equation}
A_{x,y} = \exp\left( -\beta d_{x,y} \right)
\end{equation}

where $d_{x,y}$ is the distance to the nearest known pixel, computed as:

\begin{equation}
d_{x,y} = \min_{(i,j) \in \mathcal{K}} \sqrt{(x - i)^2 + (y - j)^2}
\end{equation}

and $\beta$ is a scaling factor controlling the rate of decay, and $\mathcal{K} = \{ (i, j) \mid M(i, j) = 0 \}$.

\begin{figure}[htbp]
\centering
\includegraphics[width=0.8\linewidth]{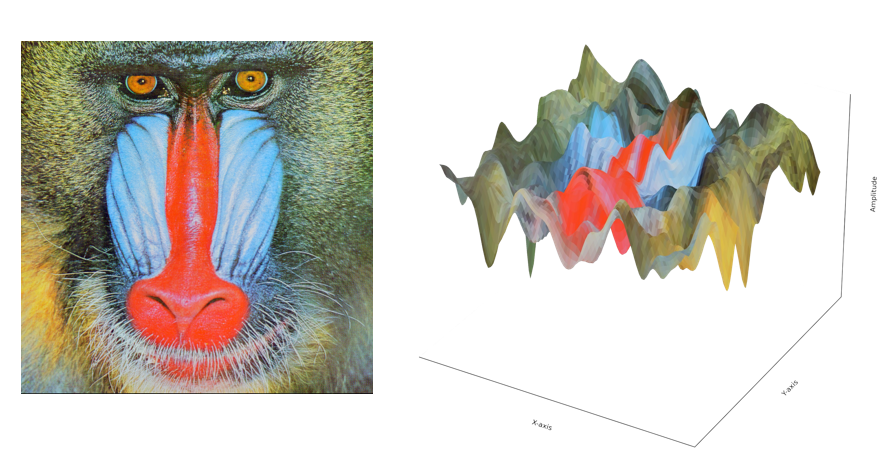}
\caption{Visualization of the Gaussian splat map with adaptive amplitudes. The left image shows the original image, and the right image displays the corresponding 3D representation of the Gaussian splat map.}
\label{fig:amplitude}
\end{figure}

An example of the Gaussian splat map with adaptive amplitudes is illustrated in Figure~\ref{fig:amplitude}.

\subsubsection{Generation of Gaussian Splat Map}

For the entire missing region, we aggregate the anisotropic Gaussian functions weighted by their amplitudes to form the Gaussian splat map $S$:

\begin{equation}
S(u, v) = \sum_{(x, y) \in \mathcal{M}} A_{x,y} \, G_{x,y}(u, v)
\end{equation}

where $\mathcal{M} = \{ (x, y) \mid M(x, y) = 1 \}$.

\subsubsection{Normalization and Multi-Scale Integration}

To capture information at different scales, we compute the Gaussian splat maps at multiple scales by varying the neighborhood size in covariance estimation~\cite{lindeberg1994scale}. The final splat map $S_{\text{norm}}$ is obtained by normalizing and combining the multi-scale splat maps:

\begin{equation}
S_{\text{norm}}(u, v) = \frac{S(u, v) - S_{\text{min}}}{S_{\text{max}} - S_{\text{min}}}
\end{equation}

where $S_{\text{min}}$ and $S_{\text{max}}$ are the minimum and maximum values of $S$ across all scales.

\subsection{Diffusion-Based Inpainting Network with Anisotropic Gaussian Splatting}

We integrate the anisotropic Gaussian splat maps into a diffusion-based inpainting network to guide the restoration of missing regions. The diffusion model leverages the spatial influence modeled by the Gaussian splats to generate more accurate and coherent inpainting results.

\subsubsection{Diffusion Process for Image Inpainting}

We employ a diffusion model for image inpainting, where the model learns to reverse a diffusion process that gradually adds noise to the image~\cite{sohl2015deep, ho2020denoising}. In our context, the diffusion model is conditioned on the incomplete image and the guidance provided by the anisotropic Gaussian splat maps.

\subsubsection{Network Architecture}

Instead of utilizing a UNet architecture, we design a simple convolutional neural network (CNN) as the backbone of our diffusion model. The CNN consists of encoder and decoder layers without skip connections, processing the input data to predict the added noise at each timestep.

\subsubsection{Input to the Network}

The input to the network $\mathcal{N}$ is a concatenation of the incomplete image, the mask, the Gaussian splat map, the edge map, and the attention map derived from the splat map:

\begin{equation}
X = \left[ (I \odot (1 - M)),\ M,\ S_{\text{norm}},\ E,\ A_{\text{att}} \right]
\end{equation}

where $I$ is the original image, $M$ is the binary mask indicating missing regions, $S_{\text{norm}}$ is the normalized Gaussian splat map, $E$ is the edge map, and $A_{\text{att}}$ is the attention map computed from $S_{\text{norm}}$:

\begin{equation}
A_{\text{att}} = \sigma\left( \text{Conv}(S_{\text{norm}}) \right)
\end{equation}

with $\sigma$ being the sigmoid activation function and $\text{Conv}$ representing a convolutional operation.

\begin{figure}[htbp]
\centering
\includegraphics[width=0.9\linewidth]{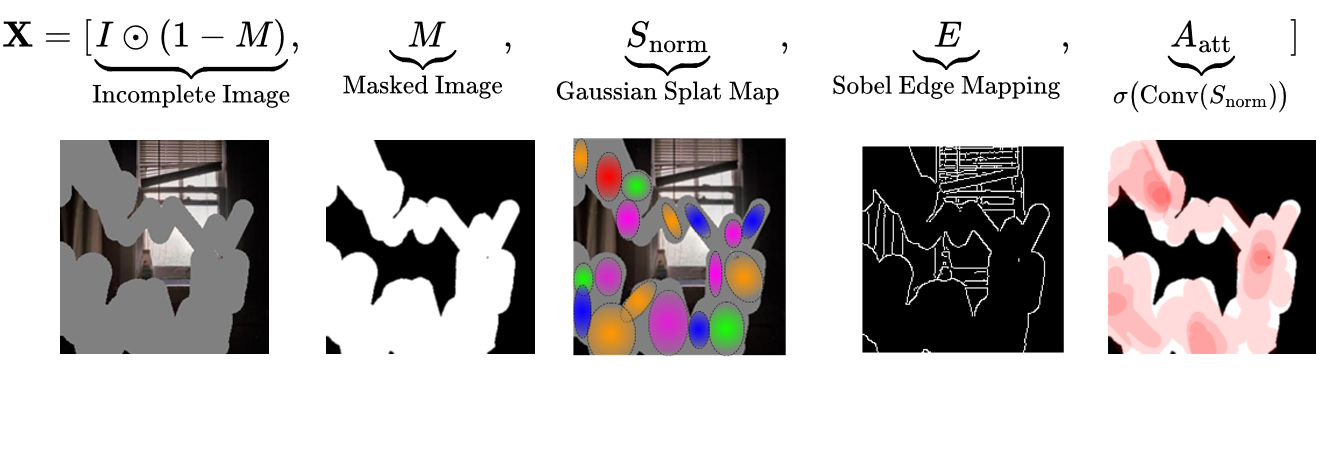}
\caption{The inputs to the inpainting network, including the incomplete image $(I \odot (1 - M))$, the mask $M$, the normalized Gaussian splat map $S_{\text{norm}}$, the edge map $E$, and the attention map $A_{\text{att}}$.}
\label{fig:input}
\end{figure}

An illustration of the inputs to the network is shown in Figure~\ref{fig:input}.

\subsubsection{Diffusion Model Training}

During training, we simulate the forward diffusion process by adding Gaussian noise to the complete image $I$ at various timesteps $t$, according to a predefined noise schedule~\cite{ho2020denoising}:

\begin{equation}
\tilde{I}_t = \sqrt{\alpha_t} I + \sqrt{1 - \alpha_t} \epsilon
\end{equation}

where $\alpha_t$ are predefined constants, and $\epsilon \sim \mathcal{N}(0, \mathbf{I})$ is standard Gaussian noise.

The network $\mathcal{N}$ is trained to predict the added noise $\epsilon$ given the noisy image $\tilde{I}_t$ and the input $X$:

\begin{equation}
\hat{\epsilon} = \mathcal{N}(X, t)
\end{equation}

\subsubsection{Loss Functions}

To train the diffusion model, we use a combination of loss functions that encourage accurate noise prediction and promote perceptually plausible inpainting results.

\paragraph{Noise Prediction Loss ($\mathcal{L}_{\text{noise}}$):}

\begin{equation}
\mathcal{L}_{\text{noise}} = \mathbb{E}_{I, t, \epsilon} \left[ \| \hat{\epsilon} - \epsilon \|_2^2 \odot (1 - M) \right]
\end{equation}

This loss encourages the network to predict the noise added at each timestep, focusing on the masked regions.

\paragraph{Reconstruction Loss ($\mathcal{L}_{\text{rec}}$):}

\begin{equation}
\mathcal{L}_{\text{rec}} = \| (\hat{I} - I) \odot (1 - M) \|_1
\end{equation}

where $\hat{I}$ is the reconstructed image obtained after the reverse diffusion process.

\paragraph{Perceptual Loss ($\mathcal{L}_{\text{perc}}$):}

\begin{equation}
\mathcal{L}_{\text{perc}} = \sum_{l} \| \phi_l(\hat{I}) - \phi_l(I) \|_1
\end{equation}

where $\phi_l$ represents the activation maps from layer $l$ of a pretrained VGG-19 network~\cite{simonyan2014very}.

\paragraph{Style Loss ($\mathcal{L}_{\text{style}}$):}

\begin{equation}
\mathcal{L}_{\text{style}} = \sum_{l} \| \mathcal{G} \left( \phi_l(\hat{I}) \right) - \mathcal{G} \left( \phi_l(I) \right) \|_1
\end{equation}

where $\mathcal{G}(\cdot)$ computes the Gram matrix of the feature maps~\cite{gatys2016image}.

\paragraph{Total Variation Loss ($\mathcal{L}_{\text{tv}}$):}

\begin{equation}
\mathcal{L}_{\text{tv}} = \sum_{i,j} \left( (\hat{I}_{i+1,j} - \hat{I}_{i,j})^2 + (\hat{I}_{i,j+1} - \hat{I}_{i,j})^2 \right)
\end{equation}

\paragraph{Total Loss:}

The total loss function is a weighted sum of the individual losses:

\begin{equation}
\mathcal{L}_{\text{total}} = \lambda_{\text{noise}} \mathcal{L}_{\text{noise}} + \lambda_{\text{rec}} \mathcal{L}_{\text{rec}} + \lambda_{\text{perc}} \mathcal{L}_{\text{perc}} + \lambda_{\text{style}} \mathcal{L}_{\text{style}} + \lambda_{\text{tv}} \mathcal{L}_{\text{tv}}
\end{equation}

where $\lambda_{\text{noise}}$, $\lambda_{\text{rec}}$, $\lambda_{\text{perc}}$, $\lambda_{\text{style}}$, and $\lambda_{\text{tv}}$ are hyperparameters.

\vspace{2mm}
\noindent
{\small Our code is available at \url{https://github.com/jacobfa/splatting}.}

\section{Experiments}

In this section, we evaluate the proposed image inpainting method on two benchmark datasets: CIFAR-10 and CelebA. We provide details about the datasets, explain the inpainting and masking processes, describe the implementation details, and present the evaluation metrics used to assess the performance.

\subsection{Datasets}

\subsubsection{CIFAR-10}

The CIFAR-10 dataset~\cite{krizhevsky2009learning} consists of $60{,}000$ color images in $10$ classes, with $6{,}000$ images per class. Each image has a spatial resolution of $32 \times 32$ pixels. We use the standard split of $50{,}000$ images for training and $10{,}000$ images for testing.

\subsubsection{CelebA}

The CelebFaces Attributes Dataset (CelebA)~\cite{liu2015faceattributes} contains over $200{,}000$ celebrity images, each annotated with $40$ attributes. The images exhibit large variations in pose, facial expression, and background. We preprocess the images by cropping and resizing them to $64 \times 64$ pixels to accommodate our model's input requirements.

\subsection{Inpainting and Masking Process}

\subsubsection{Inpainting Process Overview}

Image inpainting aims to fill missing regions in images in a visually plausible way. Our experiments simulate missing regions by applying randomly generated masks to the images. The inpainting process involves using incomplete images and the corresponding masks as inputs to our diffusion-based network, which predicts the missing content conditioned on the known regions.

\subsubsection{Mask Generation}

We generate irregular masks to mimic natural occlusions or corruptions in images. The mask-generation process includes the following steps:

\begin{itemize}
    \item \textbf{Random Brush Strokes}: We create masks using random brush strokes with varying widths, directions, and lengths. This simulates realistic occlusions.
    \item \textbf{Coverage Control}: We adjust the number and size of brush strokes to cover approximately $20\%$ of the image area, ensuring sufficient challenge for the inpainting task.
    \item \textbf{Binary Mask Representation}: The generated masks are binary images where pixels in the known regions have a value of $1$, and pixels in the missing regions have a value of $0$.
\end{itemize}

Mathematically, the mask $M(x, y)$ for pixel location $(x, y)$ is defined as:

\begin{equation}
M(x, y) =
\begin{cases}
1, & \text{if } (x, y) \text{ is in the known region} \\
0, & \text{if } (x, y) \text{ is in the missing region}
\end{cases}
\end{equation}

\subsubsection{Incomplete Image Creation}

Given an original image $I$ and its corresponding mask $M$, we generate the incomplete image $I_{\text{incomplete}}$ by element-wise multiplication:

\begin{equation}
I_{\text{incomplete}} = I \odot M
\end{equation}

where $\odot$ denotes the Hadamard (element-wise) product, this operation retains the pixel values in the known regions and sets the pixel values in the missing regions to zero.

\subsection{Implementation Details}

\subsubsection{Network Architecture}

We implement the diffusion-based inpainting network as described in Section~3.2. The network uses a simplified convolutional neural network (CNN) as the backbone, processing inputs without skip connections. The input to the network includes:

\begin{itemize}
    \item \textbf{Incomplete Image} $(I \odot M)$: The partially observed image.
    \item \textbf{Mask} $M$: Indicates the locations of known and missing regions.
    \item \textbf{Gaussian Splat Map} $S_{\text{norm}}$: Encodes spatial influence around missing regions.
    \item \textbf{Edge Map} $E$: Captures structural information of the image.
    \item \textbf{Attention Map} $A_{\text{att}}$: Derived from the Gaussian splat map to guide the network's focus.
\end{itemize}

These components are concatenated along the channel dimension and fed into the network.

\subsubsection{Training Parameters}

We train our model using the following settings:

\begin{itemize}
    \item \textbf{Optimizer}: AdamW~\cite{loshchilov2017decoupled} with $\beta_1 = 0.9$, $\beta_2 = 0.999$.
    \item \textbf{Learning Rate}: $2 \times 10^{-4}$.
    \item \textbf{Batch Size}: $128$.
    \item \textbf{Number of Epochs}: $200$.
    \item \textbf{Loss Weights}:
    \begin{itemize}
        \item $\lambda_{\text{noise}} = 1.0$
        \item $\lambda_{\text{rec}} = 5.0$
        \item $\lambda_{\text{perc}} = 0.5$
        \item $\lambda_{\text{style}} = 1.0$
        \item $\lambda_{\text{tv}} = 0.05$
    \end{itemize}
\end{itemize}

\subsubsection{Hyperparameters}

The hyperparameters used in our experiments are:

\begin{itemize}
    \item \textbf{Covariance Regularization Constant} ($\epsilon$): $1 \times 10^{-5}$.
    \item \textbf{Amplitude Decay Factor} ($\beta$): $0.1$.
\end{itemize}

\subsubsection{Edge Map Computation}

We compute the edge maps $E$ using the Sobel operator~\cite{kanopoulos1988design}. The gradients in the $x$ and $y$ directions are computed as:

\begin{equation}
E_x = I * S_x, \quad E_y = I * S_y
\end{equation}

where $S_x$ and $S_y$ are the Sobel kernels. The magnitude of the gradient is then:

\begin{equation}
E = \sqrt{E_x^2 + E_y^2}
\end{equation}

\subsubsection{Gaussian Splat Map Computation}

Using the edge maps, we estimate the covariance matrices $\Sigma_{x,y}$ as described in Section~3.1.2. We compute the amplitude map $A_{x,y}$ based on the distance to the nearest known pixel (Section~3.1.3). The Gaussian splat map $S$ is then generated by aggregating the anisotropic Gaussian functions (Section~3.1.4) and normalized across multiple scales (Section~3.1.5).

\subsection{Evaluation Metrics}

We assess the performance of our method using the following quantitative metrics:

\begin{itemize}
    \item \textbf{Mean Squared Error (MSE)}: Measures the average squared difference between the reconstructed image $\hat{I}$ and the ground truth image $I$.

    \begin{equation}
    \text{MSE} = \frac{1}{N} \sum_{i=1}^{N} \left( \hat{I}_i - I_i \right)^2
    \end{equation}

    \item \textbf{Peak Signal-to-Noise Ratio (PSNR)}: Evaluates the reconstruction quality by comparing the maximum possible signal to the noise level.

    \begin{equation}
    \text{PSNR} = 20 \log_{10} \left( \frac{I_{\text{max}}}{\sqrt{\text{MSE}}} \right)
    \end{equation}

    where $I_{\text{max}}$ is the maximum possible pixel value.

    \item \textbf{Structural Similarity Index Measure (SSIM)}~\cite{wang2004image}: Assesses the structural similarity between $\hat{I}$ and $I$.

    \begin{equation}
    \text{SSIM} (\hat{I}, I) = \frac{ (2 \mu_{\hat{I}} \mu_{I} + c_1) (2 \sigma_{\hat{I} I} + c_2) }{ (\mu_{\hat{I}}^2 + \mu_{I}^2 + c_1)(\sigma_{\hat{I}}^2 + \sigma_{I}^2 + c_2) }
    \end{equation}

    where $\mu$ and $\sigma$ denote the mean and standard deviation, and $c_1$, $c_2$ are constants.
\end{itemize}

\subsection{Results}

\subsubsection{Quantitative Evaluation}

We report the averaged MSE, PSNR, and SSIM metrics on the test sets of CIFAR-10 and CelebA. Our method demonstrates superior performance compared to baseline methods, particularly regarding PSNR and SSIM, indicating higher fidelity and structural preservation in the inpainted images.

\begin{table}[h]
    \centering
    \begin{tabular}{lccc}
        \toprule
        \textbf{Method} & 
        \textbf{MSE} \textcolor{green}{$\downarrow$} & 
        \textbf{PSNR (dB)} \textcolor{green}{$\uparrow$} & 
        \textbf{SSIM} \textcolor{green}{$\uparrow$} \\
        \midrule
        Contextual Attention~\cite{yu2018generative} & $9.28 \times 10^{-3}$ & 29.87 & 0.9644 \\
        EdgeConnect~\cite{nazeri2019edgeconnect} & $5.96 \times 10^{-3}$ & 31.37 & 0.9733 \\
        Partial Convolution~\cite{liu2018image} & $7.44 \times 10^{-3}$ & 30.04 & 0.9642 \\
        \textbf{Ours} & $\bf{1.32 \times 10^{-3}}$ & \textbf{34.98} & \textbf{0.9923} \\
        \bottomrule
    \end{tabular}
    \vspace{10pt} 
    \caption{Quantitative comparison of inpainting methods on the CIFAR-10 dataset.}
    \label{tab:quantitative_results}
\end{table}

\subsubsection{Qualitative Evaluation}

\begin{figure*}[h]
\centering
\includegraphics[scale=0.3]{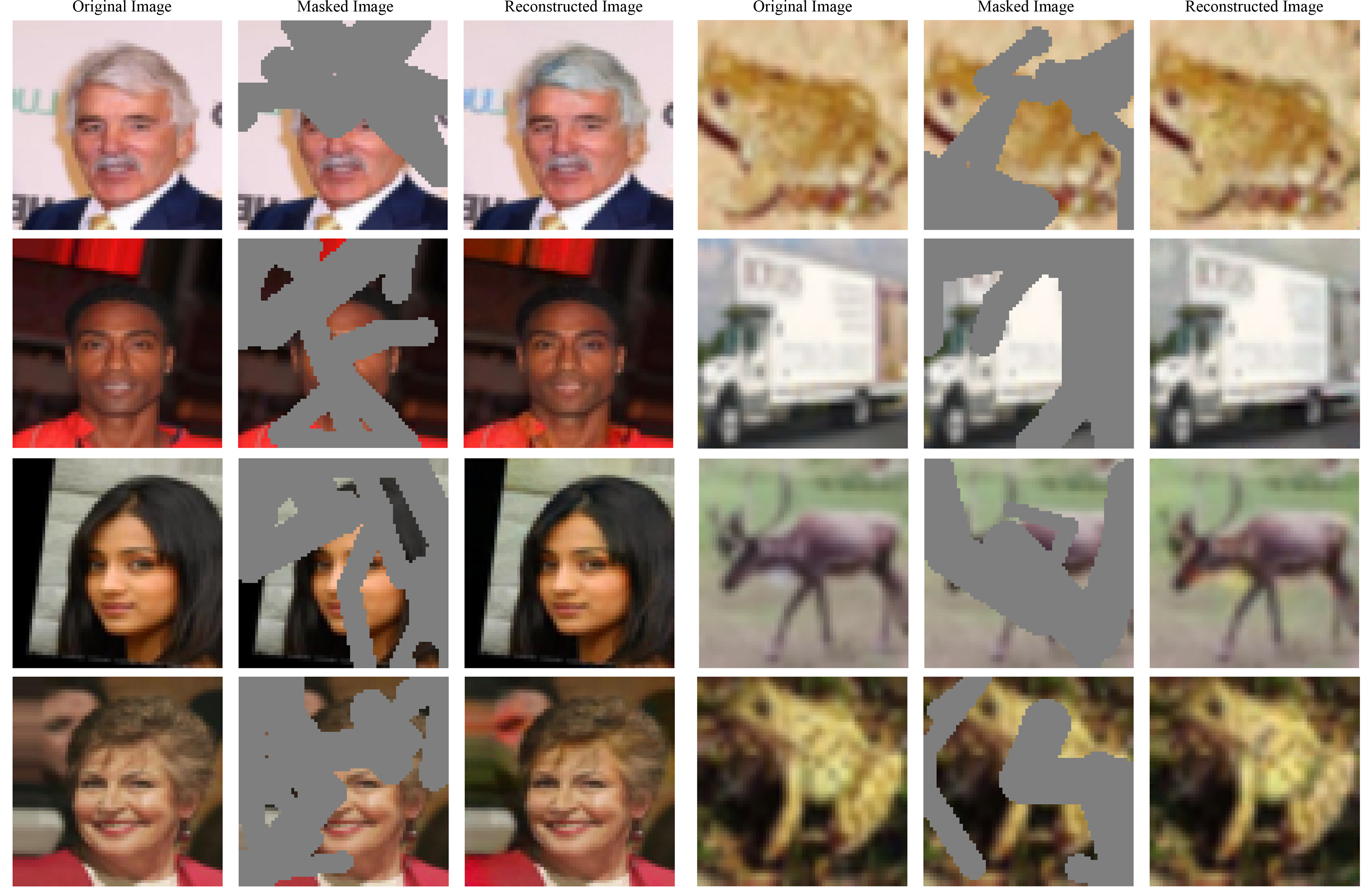}
\caption{Qualitative inpainting results on the CelebA dataset (left) and CIFAR-10 dataset (right)}
\label{fig:qualitative_results}
\end{figure*}

Figure~\ref{fig:qualitative_results} shows example inpainting results on the CelebA dataset. Our method effectively reconstructs the missing regions with realistic textures and seamless blending with the known regions.

\subsection{Comparative Experiments}

We compare our method with state-of-the-art inpainting algorithms:

\begin{itemize}
    \item \textbf{Contextual Attention Network}~\cite{yu2018generative}: This method uses a contextual attention mechanism to borrow relevant patches from known regions.
    \item \textbf{EdgeConnect}~\cite{nazeri2019edgeconnect}: An edge-guided approach that predicts edges and uses them to guide the inpainting process.
    \item \textbf{Partial Convolutional Network}~\cite{liu2018image}: Employs partial convolutions that condition the convolution on the validity of the input pixels.
    \item \textbf{Diffusion-Based Inpainting}~\cite{lugmayr2022repaint}: Utilizes diffusion models for image reconstruction.
\end{itemize}

Our method outperforms these baseline methods in quantitative metrics and visual quality, demonstrating the effectiveness of the anisotropic Gaussian splatting and the diffusion-based inpainting network.

\subsection{Ablation Studies}

To assess the contributions of different components, we perform ablation studies by removing or modifying parts of our model:

\begin{itemize}
    \item \textbf{Without Gaussian Splatting}: Removing the Gaussian splatting module leads to less accurate structure reconstruction.
    \item \textbf{Without Edge Guidance}: Excluding the edge maps from the input degrades performance in preserving image edges.
    \item \textbf{Alternative Loss Functions}: Replacing the perceptual and style losses with simple pixel-wise losses results in less visually pleasing outputs.
\end{itemize}

\subsection{Discussion}

Our experiments validate the proposed method's capability to reconstruct missing regions with high fidelity. The anisotropic Gaussian splatting effectively captures spatial influence around missing regions, and the diffusion-based network leverages this information for plausible inpainting. The combination of loss functions contributes to preserving both global structures and fine details.

\section{Conclusion}

We have presented a novel image inpainting method that integrates diffusion models with anisotropic Gaussian splatting. Through comprehensive experiments on CIFAR-10 and CelebA datasets, we have demonstrated the effectiveness of our approach in producing high-quality inpainted images. Future work includes extending the method to higher-resolution images and exploring applications in video inpainting.

\bibliographystyle{ieeetr}
\bibliography{bib}

\begin{thebibliography}{10}

\bibitem{bertalmio2000image}
M.~Bertalmio, G.~Sapiro, V.~Caselles, and C.~Ballester, ``Image inpainting,'' in {\em Proceedings of the 27th Annual Conference on Computer Graphics and Interactive Techniques}, pp.~417--424, ACM, 2000.

\bibitem{criminisi2004region}
A.~Criminisi, P.~P{\'e}rez, and K.~Toyama, ``Region filling and object removal by exemplar-based image inpainting,'' {\em IEEE Transactions on Image Processing}, vol.~13, no.~9, pp.~1200--1212, 2004.

\bibitem{ballester2001filling}
C.~Ballester, M.~Bertalmio, V.~Caselles, G.~Sapiro, and J.~Verdera, ``A variational model for filling-in gray level and color images,'' {\em IEEE Transactions on Image Processing}, vol.~10, no.~8, pp.~1200--1211, 2001.

\bibitem{pathak2016context}
D.~Pathak, P.~Kr{\"a}henb{\"u}hl, J.~Donahue, T.~Darrell, and A.~A. Efros, ``Context encoders: Feature learning by inpainting,'' in {\em Proceedings of the IEEE Conference on Computer Vision and Pattern Recognition}, pp.~2536--2544, 2016.

\bibitem{iizuka2017globally}
S.~Iizuka, E.~Simo-Serra, and H.~Ishikawa, ``Globally and locally consistent image completion,'' in {\em ACM Transactions on Graphics}, vol.~36, pp.~1--14, ACM, 2017.

\bibitem{yu2018generative}
J.~Yu, Z.~Lin, J.~Yang, X.~Shen, X.~Lu, and T.~S. Huang, ``Generative image inpainting with contextual attention,'' in {\em Proceedings of the IEEE Conference on Computer Vision and Pattern Recognition}, pp.~5505--5514, 2018.

\bibitem{liu2018image}
G.~Liu, F.~A. Reda, K.~J. Shih, T.-C. Wang, A.~Tao, and B.~Catanzaro, ``Image inpainting for irregular holes using partial convolutions,'' in {\em Proceedings of the European Conference on Computer Vision}, pp.~85--100, 2018.

\bibitem{yu2019free}
J.~Yu, Z.~Lin, J.~Yang, X.~Shen, X.~Lu, and T.~Huang, ``Free-form image inpainting with gated convolution,'' in {\em Proceedings of the IEEE/CVF International Conference on Computer Vision}, pp.~4471--4480, 2019.

\bibitem{sohl2015deep}
J.~Sohl-Dickstein, E.~A. Weiss, N.~Maheswaranathan, and S.~Ganguli, ``Deep unsupervised learning using nonequilibrium thermodynamics,'' 2015.

\bibitem{ho2020denoising}
J.~Ho, A.~Jain, and P.~Abbeel, ``Denoising diffusion probabilistic models,'' 2020.

\bibitem{song2020score}
Y.~Song, J.~Sohl-Dickstein, D.~P. Kingma, A.~Kumar, S.~Ermon, and B.~Poole, ``Score-based generative modeling through stochastic differential equations,'' 2021.

\bibitem{gao2023surfelnerfneuralsurfelradiance}
Y.~Gao, Y.-P. Cao, and Y.~Shan, ``Surfelnerf: Neural surfel radiance fields for online photorealistic reconstruction of indoor scenes,'' 2023.

\bibitem{efros1999texture}
A.~A. Efros and T.~K. Leung, ``Texture synthesis by non-parametric sampling,'' in {\em Proceedings of the Seventh IEEE International Conference on Computer Vision}, vol.~2, pp.~1033--1038, IEEE, 1999.

\bibitem{nazeri2019edgeconnect}
K.~Nazeri, E.~Ng, T.~Joseph, F.~Qureshi, and M.~Ebrahimi, ``Edgeconnect: Generative image inpainting with adversarial edge learning,'' {\em arXiv preprint arXiv:1901.00212}, 2019.

\bibitem{saharia2022palette}
C.~Saharia, W.~Chan, H.~Chang, C.~A. Lee, J.~Ho, T.~Salimans, D.~J. Fleet, and M.~Norouzi, ``Palette: Image-to-image diffusion models,'' 2022.

\bibitem{lugmayr2022repaint}
A.~Lugmayr, M.~Danelljan, A.~Romero, F.~Yu, R.~Timofte, and L.~V. Gool, ``Repaint: Inpainting using denoising diffusion probabilistic models,'' 2022.

\bibitem{ronneberger2015u}
O.~Ronneberger, P.~Fischer, and T.~Brox, ``U-net: Convolutional networks for biomedical image segmentation,'' in {\em International Conference on Medical Image Computing and Computer-Assisted Intervention}, pp.~234--241, Springer, 2015.

\bibitem{vaswani2017attention}
A.~Vaswani, N.~Shazeer, N.~Parmar, J.~Uszkoreit, L.~Jones, A.~N. Gomez, L.~Kaiser, and I.~Polosukhin, ``Attention is all you need,'' in {\em Advances in Neural Information Processing Systems}, pp.~5998--6008, 2017.

\bibitem{perona1990scale}
P.~Perona and J.~Malik, ``Scale-space and edge detection using anisotropic diffusion,'' {\em IEEE Transactions on Pattern Analysis and Machine Intelligence}, vol.~12, no.~7, pp.~629--639, 1990.

\bibitem{canny1986computational}
J.~Canny, ``A computational approach to edge detection,'' {\em IEEE Transactions on Pattern Analysis and Machine Intelligence}, vol.~PAMI-8, no.~6, pp.~679--698, 1986.

\bibitem{lindeberg1994scale}
T.~Lindeberg, {\em Scale-space theory in computer vision}, vol.~256.
\newblock Springer Science \& Business Media, 1994.

\bibitem{simonyan2014very}
K.~Simonyan and A.~Zisserman, ``Very deep convolutional networks for large-scale image recognition,'' {\em arXiv preprint arXiv:1409.1556}, 2014.

\bibitem{gatys2016image}
L.~A. Gatys, A.~S. Ecker, and M.~Bethge, ``Image style transfer using convolutional neural networks,'' {\em Proceedings of the IEEE Conference on Computer Vision and Pattern Recognition}, pp.~2414--2423, 2016.

\bibitem{krizhevsky2009learning}
A.~Krizhevsky, ``Learning multiple layers of features from tiny images,'' tech. rep., Technical report, University of Toronto, 2009.

\bibitem{liu2015faceattributes}
Z.~Liu, P.~Luo, X.~Wang, and X.~Tang, ``Deep learning face attributes in the wild,'' in {\em Proceedings of the IEEE International Conference on Computer Vision (ICCV)}, pp.~3730--3738, 2015.

\bibitem{loshchilov2017decoupled}
I.~Loshchilov and F.~Hutter, ``Decoupled weight decay regularization,'' {\em arXiv preprint arXiv:1711.05101}, 2017.

\bibitem{kanopoulos1988design}
N.~Kanopoulos, N.~Vasanthavada, and R.~L. Baker, ``Design of an image edge detection filter using the sobel operator,'' {\em IEEE Journal of Solid-State Circuits}, vol.~23, no.~2, pp.~358--367, 1988.

\bibitem{wang2004image}
Z.~Wang, A.~C. Bovik, H.~R. Sheikh, and E.~P. Simoncelli, ``Image quality assessment: From error visibility to structural similarity,'' {\em IEEE Transactions on Image Processing}, vol.~13, no.~4, pp.~600--612, 2004.

\end{thebibliography}

\end{document}